\documentclass[10pt,twocolumn,letterpaper]{article}

\usepackage{cvpr}
\usepackage{times}
\usepackage{epsfig}
\usepackage{graphicx}
\usepackage{amsmath}
\usepackage{amssymb}

\usepackage[ruled,vlined,commentsnumbered]{algorithm2e}
\usepackage{url}

\newcommand{\qed}{\nobreak \ifvmode \relax \else
      \ifdim\lastskip<1.5em \hskip-\lastskip
      \hskip1.5em plus0em minus0.5em \fi \nobreak
      \vrule height0.75em width0.5em depth0.25em\fi}
\usepackage[pagebackref=true,breaklinks=true,letterpaper=true,colorlinks,bookmarks=false]{hyperref}

\cvprfinalcopy 

\def\cvprPaperID{231} 

\ifcvprfinal\fi
\begin{document}

\title{ Unsupervised Object Discovery and Localization in the Wild: \\Part-based Matching with Bottom-up Region Proposals }

\author{Minsu Cho\textsuperscript{1,}\thanks{WILLOW project-team, D\'epartement d'Informatique de l'Ecole Normale
Sup\'erieure, ENS/Inria/CNRS UMR 8548.} \quad\quad\quad\quad Suha Kwak\textsuperscript{1,}\footnotemark[1] \quad\quad\quad\quad Cordelia Schmid\textsuperscript{1,}\thanks{LEAR project-team, Inria Grenoble Rh\^one-Alpes, LJK, CNRS, Univ. Grenoble Alpes, France.}  \quad\quad\quad Jean Ponce\textsuperscript{2,}\footnotemark[1] \vspace*{0.2cm}\\
{\textsuperscript{1}Inria \quad\quad\quad\quad\quad \textsuperscript{2}\'{E}cole Normale Sup\'erieure / PSL Research University} \vspace*{0.2cm}\\
}

\maketitle
\thispagestyle{empty}

\begin{abstract}
This paper addresses unsupervised discovery and localization of
dominant objects from a noisy image collection with multiple object
classes. The setting of this problem is fully unsupervised, without
even image-level annotations or any assumption of a single dominant
class.  This is far more general than typical
colocalization, cosegmentation, or weakly-supervised localization
tasks.  We tackle the discovery and localization problem using a
part-based region matching approach: We use off-the-shelf region 
proposals to form a set of candidate bounding boxes for objects and
object parts. These regions are efficiently matched across images
using a probabilistic Hough transform that evaluates the confidence for 
each candidate correspondence considering both appearance and spatial
consistency. Dominant objects are discovered and localized by comparing the scores of candidate regions and selecting those that stand out over other regions containing them. Extensive experimental evaluations on standard benchmarks
demonstrate that the proposed approach significantly outperforms the current state of the art in
colocalization, and achieves robust object discovery in challenging mixed-class datasets.
\end{abstract}\vspace{-0.4cm}
\section{Introduction}
Object localization and detection is highly challenging because of intra-class variations, background clutter, and occlusions present in real-world images. 
While significant progress has been made in this area over the last decade, as shown by recent benchmark results~\cite{imagenet_cvpr09,pascal-voc-2007}, most state-of-the-art methods still rely on strong supervision in the form of manually-annotated bounding boxes on target instances. 
Since those detailed annotations are expensive to acquire and also prone to unwanted biases and errors, 
recent work has explored the problem of weakly-supervised object discovery where instances of an object class are found in a collection of images without any box-level annotations. Typically, weakly-supervised localization~\cite{Cinbis2014,Nguyen09,Pandey2011,shi2013bayesian,siva2012defence,wang2014weakly} requires positive and negative image-level labels for a target object class. 
On the other hand, cosegmentation~\cite{Joulin2010,Kim2011,Rubinstein2013} and colocalization~\cite{Deselaers:2010he,Joulin14,Tang14} assume less supervision and only require the image collection to contain a single dominant object class, allowing noisy images to some degree. 

\begin{figure}[t]
\begin{minipage}{1.0\linewidth}
\center
\includegraphics[width=0.98\linewidth]{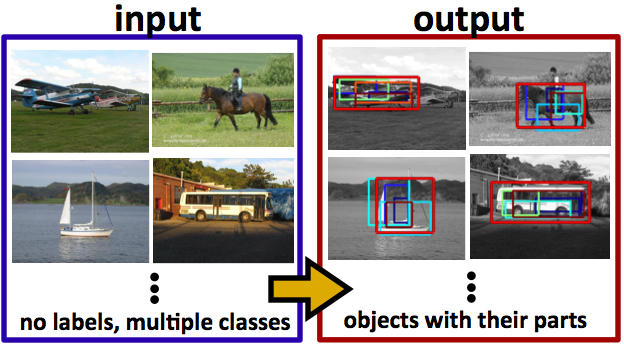}
\end{minipage}\vskip0.1cm
\caption{ Unsupervised object discovery in the wild. We tackle object localization in an unsupervised scenario without any type of annotations, where a given image collection may contain multiple dominant object classes and even outlier images. The proposed method discovers object instances (red bounding boxes) with their distinctive parts (smaller boxes). (Best viewed in color.) }
\label{fig:intro}
\vspace{-0.3cm}
\end{figure}

This paper addresses unsupervised object localization in a far more general scenario where a given image collection contain {\it multiple dominant object classes} and even {\it noisy images} without any target objects. As illustrated in Fig.~\ref{fig:intro}, the setting of this problem is fully unsupervised, without any image-level annotations, an assumption of a single dominant
class, or even a known number of object classes. In spite of this generality, the proposed method markedly outperforms the state of the arts in colocalization~\cite{Joulin14,Tang14} on standard benchmarks~\cite{pascal-voc-2007,Rubinstein2013}, and closely competes with current weakly-supervised localization~\cite{Cinbis2014,shi2013bayesian,wang2014weakly}.

We advocate a part-based matching approach to unsupervised object discovery using bottom-up region proposals. 
Multi-scale region proposals have been widely used before to restrict the search space for object bounding boxes in object recognition~\cite{Cinbis2014,girshick2014rcnn,UijlingsIJCV2013} and localization~\cite{Cinbis2014,Joulin14,Tang14,Vicente2011}. We go further and propose here to use these regions to form a set of candidate regions not only for objects, but also for object parts.
We use a probabilistic Hough transorm~\cite{Ballard81} to match those candidate regions across images, and assign them confidence scores reflecting both appearance and spatial consistency. This can be seen as an unsupervised and efficient variant of both deformable part models~\cite{felzenszwalb2010object,IJCV/FelzenszwalbH03}  and graph matching methods~\cite{Cho2013,ICCV/DuchenneJP11}.
Objects are discovered and localized by selecting the most salient regions that contain corresponding parts. To this end, we introduce a score that measures how much a region stands out over other regions containing it. The proposed algorithm alternates between part-based region matching and foreground localization, improving both over iterations. 

  The main contributions of this paper can be summarized as follows: 
 (1) A part-based region matching approach to unsupervised object discovery is introduced.   
 (2) An efficient and robust matching algorithm based on a probabilistic Hough transform is proposed. 
 (3) A standout score for robust foreground localization is introduced. 
 (4) Object discovery and localization in a fully unsupervised setup is explored on challenging benchmark datasets~\cite{pascal-voc-2007,Rubinstein2013}.
 
\section{ Related work}

 Unsupervised object discovery has long been attempted in computer vision. Sivic \etal~\cite{ICCV/SivicREZF05} and Russell \etal~\cite{Russell06} apply statistical topic discovery models. 
Grauman and Darrel~\cite{grauman2006unsupervised} use partial correspondence and clustering of local features.  
Kim and Torralba~\cite{kim2009unsupervised} employ a link analysis technique. 
Faktor and Irani~\cite{faktor2012clustering} propose clustering by composition.  
Unsupervised object discovery, however, has proven extremely difficult ``in the wild''; all of these previous approaches have been successfully demonstrated in a restricted setting with a few distinctive object classes,  
but their localization results turn out to be far behind weakly-supervised results on challenging benchmarks~\cite{Deselaers:2010he,kim2009unsupervised,Tang14}. 

Given the difficulty of fully unsupervised discovery, recent work has more focused on weakly-supervised approaches from different angles. Cosegmentation is the problem of segmenting common foreground regions out of a set of images. It has been first introduced by Rother \etal.~\cite{CVPR/RotherMBK06} who fuse Markov random fields with color histogram matching to segment objects common to two images. Since then, this approach has been improved in numerous ways~\cite{batra2010icoseg,ECCV/ChoSL08,hochbaum2009efficient,vicente2010cosegmentation},  
and extended to handle more general cases~\cite{CVPR/ChoSL10,Joulin2010,Rubinstein2013,Vicente2011}. 
Given the same type of input as cosegmentation, colocalization seeks to localize objects with bounding boxes instead of pixel-wise segmentations. 
Tang \etal.~\cite{Tang14} use the discriminative clustering framework of~\cite{Joulin2010} to localize common objects in a set of noisy images, and Joulin \etal.~\cite{Joulin14} extend it to colocalization of video frames. Weakly-supervised localization~\cite{Cinbis2014,Deselaers:2010he,Nguyen09,Pandey2011,Siva2013,song2014learning} shares the same type of output as colocalization, but assumes a more supervised scenario with image-level labels that indicate whether a target object class appears in the image or not. These labels enable to learn more discriminative localization methods, \eg, by mining negative images~\cite{Cinbis2014}. 
Recent work on discriminative patch discovery~\cite{endres2013learning,Singh2012,Sun2013} learns mid-level visual representations in a weakly-supervised mode, and use them for object recognition~\cite{endres2013learning,Singh2012} and discovery~\cite{doersch2014context,Sun2013}.

Region proposals have been used in many of the methods discussed so far, but most of them~\cite{Deselaers:2010he,Joulin14,kim2009unsupervised,Russell06,Tang14,Vicente2011} use relatively a small number of the best proposals (typically, less than 100 for each image) to form whole object hypotheses, often together with generic objectness measures~\cite{alexe2012measuring}. In contrast, we use a large number of region proposals (typically, between 1000 and 4000) as primitive elements for matching without any objectness priors. 
While many other approaches~\cite{CVPR/ChoSL10,Rubinstein2013,rubio2012unsupervised} also use correspondences between image pairs to discover object regions, they do not use an efficient part-based matching approach such as ours. 
Many of them~\cite{CVPR/ChoSL10,grauman2006unsupervised,Rubinstein2013} are driven by correspondence techniques, \eg, the SIFT flow~\cite{Liu2011}, based on generic local regions. In the sense that semi-local or mid-level parts are crucial for representing generic objects~\cite{felzenszwalb2010object,lazebnik2004semi}, we believe segment-level regions are more adequate for object matching and discovery.   
The work of Rubio \etal.~\cite{rubio2012unsupervised} introduces such a segment-level matching term in their cosegmentation formulation. Unlike ours, however, it requires a reasonable initialization by an objectness measure~\cite{alexe2012measuring}, and does not scale well with a large number of segments and images. 

\section{ Proposed approach}

For unsupervised object discovery, we combine an efficient part-based matching technique with a foreground localization scheme. 
In this section we first introduce the two main components of our approach, and then describe the overall algorithm for unsupervised object discovery. 

\subsection{ Part-based region matching}
\label{sec:prog_frame} 

\begin{figure*}[t]
\centering
\begin{minipage}{1.0\linewidth}
	\begin{minipage}{0.5\linewidth}
	\begin{minipage}{1.0\linewidth}
	\includegraphics[width=0.5\linewidth]{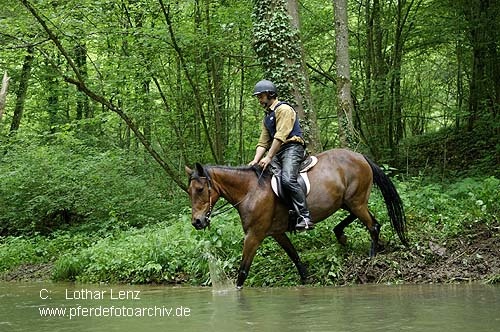}\hskip-0.0001px
	\includegraphics[width=0.5\linewidth]{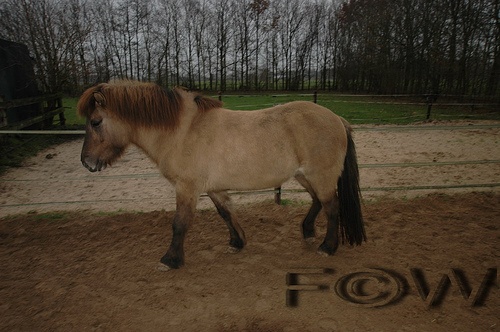}
	\end{minipage} 
	\begin{minipage}{1.0\linewidth}
	\centering
	\small{(a) Input images. }
	\end{minipage}\vskip0.1cm
	\begin{minipage}{1.0\linewidth}
	\includegraphics[width=1.0\linewidth]{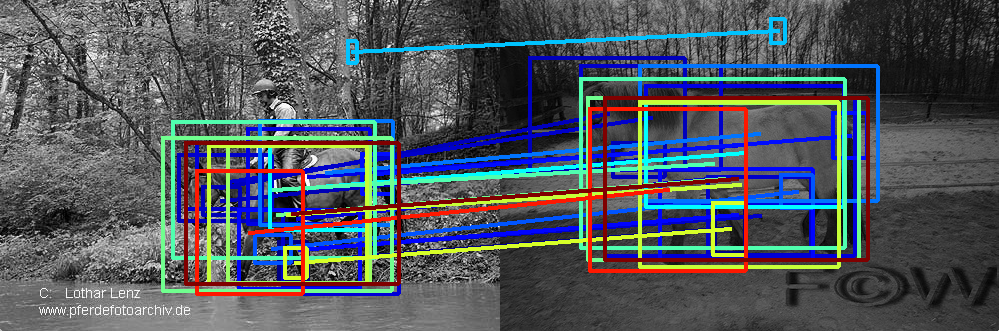}
	\end{minipage} 
	\begin{minipage}{1.0\linewidth}
	\centering
	\small{(c) Top 20 region matches.}
	\end{minipage}\vskip0.1cm
	\end{minipage}
	\begin{minipage}{0.5\linewidth}	
	\begin{minipage}{1.0\linewidth}
	\includegraphics[width=0.5\linewidth]{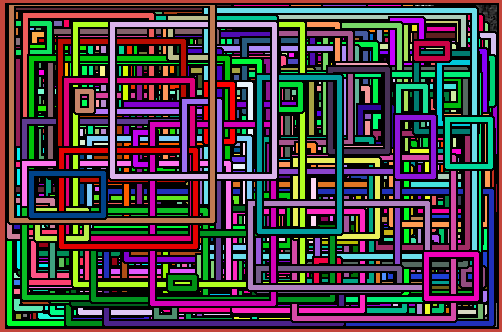}\hskip-0.0001px
	\includegraphics[width=0.5\linewidth]{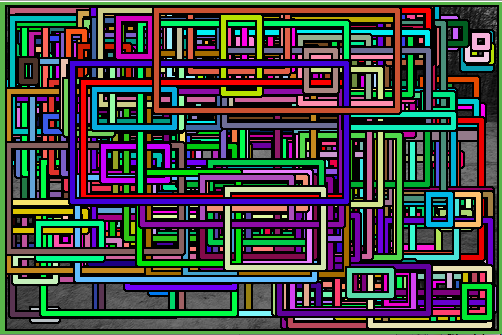}
	\end{minipage} 
	\begin{minipage}{1.0\linewidth}
	\centering
	\small{(b) Bottom-up region proposals.}
	\end{minipage}\vskip0.1cm
	\begin{minipage}{1.0\linewidth}
	\includegraphics[width=1.0\linewidth]{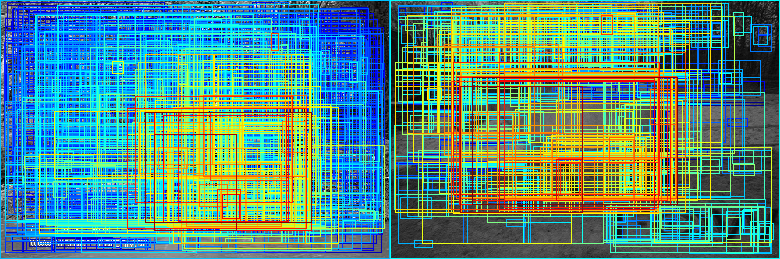}
	\end{minipage} 
	\begin{minipage}{1.0\linewidth}
	\centering
	\small{(d) Heat map representation of region confidences.}
	\end{minipage}\vskip0.1cm
	\end{minipage}
\end{minipage}\vskip0.1cm
\caption{ Part-based region matching using bottom-up region proposals. (a-b) Given two images and their multi-scale region  proposals~\cite{manen2013prime}, the proposed matching algorithm efficiently evaluates candidate matches between two sets of regions (2205$\times$1044 regions in this example) and produce match confidences for them. (c) Based on the match confidence, the 20 best matches are shown by greedy mapping with a one-to-one constraint. The confidence is color-coded in each match (red: high, blue: low). (d) The region confidences of Eq.(\ref{eq:boxconf}) are visualized in the heat map representation. Common object foregrounds tend to have higher confidences than others. (Best viewed in color.) }
\label{fig:matching}
\vspace{-0.3cm}
\end{figure*}

For part-based matching in an unsupervised setting, we use off-the-shelf region proposals~\cite{manen2013prime} as candidate regions for objects and object parts: 
Diverse multi-scale proposals include meaningful parts of objects as well as objects themselves. 
Let us assume that two sets of region proposals $R$ and $R'$ have been extracted from two images  ${\mathcal I}$ and ${\mathcal I'}$, respectively. Let $r = (f, l) \in R$ denote a region with feature $f$ observed at location $l$. 
We use $8 \times 8$ HOG descriptors~\cite{dalal2005histograms,hariharan2012discriminative} for $f$ to describe the region patches, and center position and scale for $l$ to specify the location. 
A match between $r'$ and $r$ is represented by $(r, r')$. For the sake of brevity, we use short notations $\mathcal{D}$ for two region sets, and $m$ for a match: $\mathcal{D} = ( R, R' )$, $m=(r,r')$ in $R\times R'$. 
Then, our probabilistic model of a match confidence for $m$ is represented by $p( m | \mathcal{D} )$. Assuming a common object appears in ${\mathcal I}$ and ${\mathcal I'}$, let the offset $x$ denote its pose displacement between ${\mathcal I}$ and ${\mathcal I'}$, related to properties such as position and scale change. $p(x | \mathcal{D} )$ becomes the probability of the common object being located with {\it offset} $x$.  
Now, the match confidence is decomposed in a Bayesian manner:\vspace{-0.1cm}
\begin{eqnarray}
 p( m | \mathcal{D} ) \!\!\!\!&=&\!\!\!\! \sum_{x}  p( m, x | \mathcal{D} ) =  \sum_{x}  p(  m | x, \mathcal{D} ) p(  x | \mathcal{D} ) \nonumber\\
   &=&\!\!\!\! p(  m_\text{a} ) \sum_{x}   p( m_\text{g} | x ) p( x | \mathcal{D} ), \label{eq:mconfid}
\end{eqnarray} \vskip-0.15cm \hskip-0.4cm where we suppose that appearance matching $m_\text{a}$ is independent of geometry matching $m_\text{g}$ and an object location offset $x$. Appearance likelihood $p( m_a )$ is simply computed as the similarity between $f$ and $f'$. Geometry likelihood  $p( m_\text{g} | x )$ is estimated by comparing displacement $l'-l$ to the given offset $x$. For $p( m_\text{g} | x )$, we construct three-dimensional offset bins for translation and scale change, and use a Gaussian distribution centered on offset $x$.  

The main issue is how to estimate geometry prior $p( x | \mathcal{D} )$ without any information about objects and their locations. Inspired by the generalized Hough transform~\cite{Ballard81} and its extensions~\cite{leibe2008robust,Zhang2009}, we propose the Hough space score $h(x | \mathcal{D} )$, that is the sum of individual probabilities $p(  m, x |  \mathcal{D} )$ over all possible region matches $m \in R \times R'$. 
The voting is done with an initial assumption of a uniform prior over $x$:\vspace{-0.25cm} 
\begin{eqnarray}
 h( x | \mathcal{D} )  \!\!\!\!&=&\!\!\!\! \sum_{ m } p(  m | x, \mathcal{D} ) \nonumber\\ 
 &=&\!\!\!\! \sum_{ m } p( m_\text{a} ) p( m_\text{g} | x ),\label{eq:hscore}
\end{eqnarray}
which predicts a pseudo likelihood of common objects at offset $x$. 
Assuming $p( x | \mathcal{D} ) \propto h( x | \mathcal{D} )$\footnote{Although Eq.(\ref{eq:mconfid}) is a valid probabilistic model, using $h(x | \mathcal{D} )$ as defined by Eq.(\ref{eq:hscore}) is a heuristic way of estimating $p(x|\mathcal{D})$ in terms of ``pseudo likelihood''. Like all its ``probabilistic Hough transform'' predecessors we know of~\cite{CVPR/BarinovaLK10,leibe2008robust,Maji2009}, however, it lacks a proper probabilistic interpretation.}, we rewrite Eq.(\ref{eq:mconfid}) to define the {\it Hough match confidence} as  
\begin{eqnarray}
c( m | \mathcal{D}) = p(  m_\text{a} ) \sum_{x}   p( m_\text{g}| x) h( x | \mathcal{D}). \label{eq:hmconfid}
\end{eqnarray}
Interestingly, this formulation can be seen as a combination of bottom-up and top-down processes: The bottom-up process aggregates individual votes into the Hough space scores (Eq.(\ref{eq:hscore})), and the top-down process evaluates each match confidence based on those  scores (Eq.(\ref{eq:hmconfid})). 
We call this algorithm {\it Probabilistic Hough Matching} (PHM). Leveraging the Hough space score as a spatial prior, it provides robust match confidences for candidate matches. In particular, given multi-scale region proposals, different region matches on the same object cast votes for each other, and make all the region matches on the object obtain high confidences. This is an efficient part-based matching procedure with complexity of $\mathcal{O}(nn')$, where $n$ and $n'$ are the number of regions in $R$ and $R'$, respectively. 
As shown in Fig.~\ref{fig:matching}c, reliable matches can be obtained when a proper mapping constraint (\eg, one-to-one, one-to-many, etc.) is enforced on the confidence as a post-processing.\footnote{Here we focus on the use of match confidence for object discovery rather than the final individual matches. For more details on PHM, see our webpage: \url{http://www.di.ens.fr/willow/research/PHM/}.} %

We define the {\it region confidence} as a max-pooled match confidence for $r$ in $R$ with respect to $R'$:
\begin{eqnarray}
\phi(r) = \max_{r'}{c\bigl( (r, r') | ( R, R' ) \bigr)}, \label{eq:boxconf} 
\end{eqnarray}
which derives from the best matches from $R'$ to $R$ under one-to-many mapping constraints. High region confidences guarantee that corresponding regions have at least single good matches in consideration of both appearance and spatial consistency. As shown in Fig.~\ref{fig:matching}d, the region confidence provides a useful measure for common regions between images, thus functioning as a driving force in object discovery. 

\subsection{ Foreground localization}
Foreground objects do not directly emerge from part-based region matching: 
A region with the highest confidence is often a salient part of a common object while good localization is supposed to tightly bound the entire object region.
We need a principled and unsupervised way to tackle the intrinsic ambiguity in separating the foreground objects from the background, which is one of the main challenges in unsupervised object discovery. 
In Gestalt principles of visual perception~\cite{Rubin2001} and design~\cite{jackson2008gestalt}, regions that ``stand out'' are more likely to be seen as a foreground. A high contrast lies between the foreground and background, and a lower contrast between foreground parts or background parts. 
Inspired by these figure/ground principles, we evaluate a foreground score of a region by its perceptual {\it contrast} standing out of its potential backgrounds.
To measure the contrast, we leverage on the region confidence from part-based matching, which is well supported by the work of Peterson and Gibson, demonstrating the role of object recognition or matching in the figure/ground process~\cite{peterson1994object}.
 
First, we generalize the notion of the region confidence to exploit multiple images. Let us assume $\mathcal{I}$ as a target image, and $\mathcal{I}'$ as a source image. 
The region confidence of Eq.(\ref{eq:boxconf}) is a function of region $r$ in target $R$ with its best correspondence $r'$ in source $R'$ as a latent variable. Given multiple source images, it can be naturally extended with more latent variables, meaning the best correspondences from the source images to $r$. Let us define {\it neighbor images} $N$ of target image $\mathcal{I}$ as an index set of source images where an object in $\mathcal{I}$ may appear. Generalizing Eq.(\ref{eq:boxconf}), the region confidence can be rewritten as\vskip-0.5cm
\begin{eqnarray}
\psi(r) &=&  {\max_{ \{ r'_i\}_{i \in N} }\sum_{i \in N}{c\bigl( (r, r_i') | ( R, R'_i ) \bigr)}}\nonumber \\
 &=&  \sum_{i \in N}{\max_{r'\in R'_i}{c\bigl( (r, r') | ( R, R'_i ) \bigr)}}, \label{eq:boxconf_m} 
\end{eqnarray}
which reduces to the aggregated confidence from the neighbor images. More images may give  better confidences.

\begin{figure}[t]
\centering
\begin{minipage}{1.0\linewidth}
\centering
\includegraphics[width=0.99\linewidth]{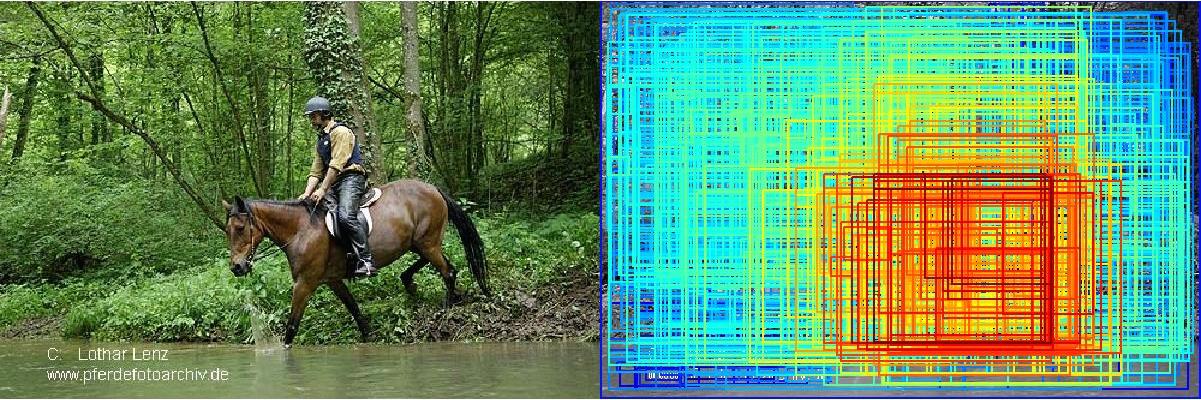}
\end{minipage}\vskip0.05cm
\begin{minipage}{1.0\linewidth}
\centering
\small{(a) Region confidences with respect to multiple source images.}
\end{minipage}\vskip0.1cm
\begin{minipage}{1.0\linewidth}
\includegraphics[width=0.5\linewidth]{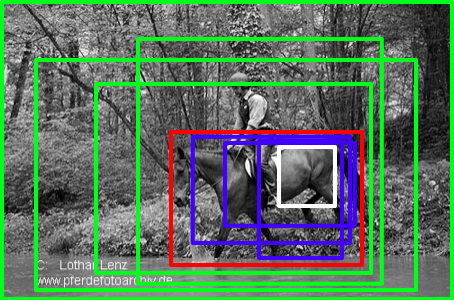}\hskip-0.0001px
\includegraphics[width=0.5\linewidth]{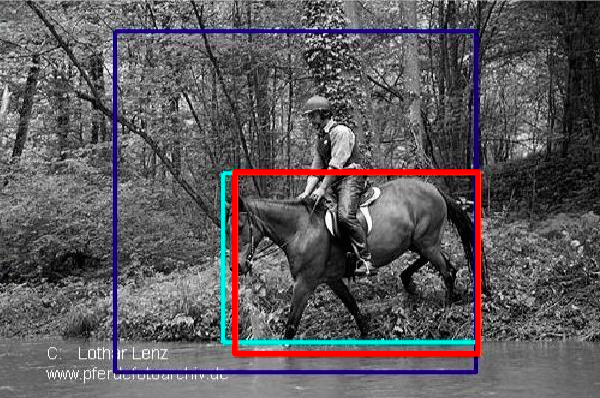}
\end{minipage}\vskip0.05cm
\begin{minipage}{1.0\linewidth}
\centering
\small{(b) Measuring the standout score from the region confidences.}
\end{minipage}\vskip0.1cm
\caption{ Foreground localization. (a) Given multiple source images with common objects, region confidences can be computed according to Eq.(\ref{eq:boxconf_m}). More source images may give better region confidences. (b) Given regions (boxes) on the left, the standout score of Eq.(\ref{eq:standout}) for the red box corresponds to the difference between its confidence and the maximum confidence of boxes containing the red box (green boxes). In the same way, the standout score for the white box takes into account blue, red, and green boxes altogether as its potential backgrounds. Three boxes on the right are ones with the top three standout scores from the region confidence. The red one has the top score. (Best viewed in color.) }
\label{fig:standout}
\vspace{-0.4cm}
\end{figure}

Given regions $R$ with their region confidences, we evaluate a perceptual contrast for region $r\in\! R$ by computing the increment of its confidence from its potential backgrounds. 
To this end, we define the {\it standout score} as\vskip-0.5cm 
\begin{eqnarray}
 s(r) = \psi(r) - \max_{r_\text{b} \in B(r)}{\psi( r_\text{b} )}, \nonumber\\
 s.t. \,\,\,\, B(r) = \{ r_\text{b} \mid r \subsetneq r_\text{b}, r_\text{b} \in R  \}, \label{eq:standout}
\end{eqnarray}
where $r \subsetneq r_\text{b}$ means region $r$ is contained in region $r_\text{b}$.  
The idea is illustrated in Fig.~\ref{fig:standout}b. 
Imagine a region gradually shrinking from a whole image region, to a tight object region, to a part region. Significant increase in region confidence is most likely to occur at the point of taking the tight object region.
In practice, we decide the inclusive relation $r \subsetneq r_\text{b}$ by two simple criteria: (1) The box area of $r$ is less than $50\%$ of the box area of $r_\text{b}$. (2) $80\%$ of the box area of $r$ overlaps with the box area of $r_\text{b}$. 

The standout score reflects the principle that we perceive a lower contrast between parts of the foreground than that between the background and the foreground. As shown in the example of Fig.~\ref{fig:standout}b, we can localize potential object regions by selecting regions with top standout scores. 

\subsection{ Object discovery algorithm}

For unsupervised object discovery, we combine part-based region matching and foreground localization in a coordinate descent-style algorithm. Given a collection of images $\mathcal{C}$, our algorithm alternates between matching image pairs and re-localizing potential object regions. Instead of matching all possible pairs over the images, we retrieve $k$ neighbors for each image and perform part-based matching only from those neighbor images. To make the algorithm robust to localization failure in precedent iterations, we maintain five potential object regions for each image. Both the neighbor images and the potential object regions are updated over iterations. 

The algorithm starts with an entire image region as an initial set of potential object regions $O_i$ for each image $\mathcal{I}_i$, and performs the following three steps at each iteration.
\vspace{-0.55cm}
\paragraph{Neighbor image retrieval.} 
For each image $\mathcal{I}_i$, $k$ nearest neighbor images $\{ \mathcal{I}_j \mid i \!\in\! N_i \}$ are retrieved based on the similarity between $O_i$ and $O_j$. We use 10 neighbor images ($k=10$).\footnote{In our experiments, the use of more neighbor images does not always improve the performance while increasing computation time.} At the first iteration, as the potential object regions become entire image regions, nearest-neighbor matching with the GIST descriptor~\cite{torralba2008small} is used. From the second iteration, we perform PHM with re-localized object regions. For efficiency, we only use the top 20 region proposals according to region confidences, which are contained in the potential object regions. The similarity for retrieval is computed as the sum of those region confidences. 
\vspace{-0.55cm}
\paragraph{Part-based region matching.} 
Part-based matching by PHM is performed on $\mathcal{I}_i$ from its neighbor images $\{ \mathcal{I}_j \mid j \!\in\! N_i \}$. To exploit current localization in a robust way, an {\it asymmetric matching strategy} is adopted: We use all regions proposals in $\mathcal{I}_i$, whereas for the neighbor image $\mathcal{I}_j$ we take regions only contained in potential object regions $O_j$. This matching strategy does not restrict  potential object regions in target $\mathcal{I}_i$ while effectively utilizing localized object regions at the precedent step.
\vspace{-0.55cm}
\paragraph{Foreground localization.} 
For each image $\mathcal{I}_i$, the standout scores are computed so that the set of potential object regions $O_i$ is updated to that of regions with top standout scores. 
 This re-localization improves both neighbor image retrieval and region matching at the subsequent iteration. 
 
 These steps are repeated for a few iterations until near-convergence. As will be shown in our experiments, 5 iterations are sufficient as no significant change occurs in more iterations. Final localization is done by selecting the most standing-out region at the end. The algorithm is designed based on the idea that better localization makes better retrieval and matching, and vice versa. As each image is independently processed at each iteration, the algorithm is  easily parallelizable in computation. Object discovery on 500 images takes less than an hour with a 10-core desktop computer, using our current parallel MATLAB implementation. 


\section{Experimental evaluation}
The degree of supervision used in visual learning tasks varies from strong (supervised localization~\cite{felzenszwalb2010object,girshick2014rcnn}) to weak (weakly-supervised localization~\cite{Cinbis2014,Siva2013}), very weak (colocalization~\cite{Joulin14,Tang14} and cosegmentation~\cite{Rubinstein2013}), and null (fully-unsupervised discovery).
To evaluate our approach for unsupervised object discovery, we conduct two types of experiments: {\it separate-class} and {\it mixed-class} experiments. Our separate-class experiments test performance of our approach in a very weakly supervised mode. Our mixed-class experiments test object discovery "in the wild" (in a fully-unsupervised mode), by mixing all images of all classes in a dataset, and evaluating performance on the whole dataset. To the best of our knowledge, 
this type of localization experiments has never been fully attempted before on challenging real-world datasets.
We conduct experiments on two realistic benchmarks, the Object Discovery~\cite{Rubinstein2013} and the P{\small ASCAL} VOC 2007~\cite{pascal-voc-2007}, and compare the results with those of the current state of the arts in cosegmentation~\cite{Kim2011,Joulin2010,Joulin2012,Rubinstein2013}, colocalization~\cite{Chum2007,Deselaers:2010he,Russell06,Joulin14,Tang14}, and weakly-supervised localization~\cite{Cinbis2014,Deselaers:2010he,Nguyen09,Pandey2011,Siva2013,wang2014weakly}. 

\subsection{Evaluation metrics}

The correct localization (CorLoc) metric is an evaluation metric widely used in related work~\cite{Deselaers:2010he,Joulin14,Siva2013,Tang14}, and defined as the percentage of images correctly localized according to the P{\small ASCAL} criterion: $\frac{area( b_p \cap b_{gt} )}{area(b_p \cup b_{gt} )} > 0.5$, where $b_p$ is the predicted box and $b_{gt}$ is the ground-truth box. The metric is adequate for a conventional separate-class setup: As a given image collection contains a single target class, only object localization is evaluated per image. 
In a mixed-class setup, however, we have another dimension involved: As different images may contain different object classes, associative relations across the images need to be evaluated.  
As such a metric orthogonal to CorLoc, we propose the {\it correct retrieval} (CorRet) evaluation metric defined as follows. Given the $k$ nearest neighbors identified by retrieval for each image, CorRet is defined as the mean percentage of these neighbors that belong to the same (ground-truth) class as the image itself.
This measure depends on $k$, fixed here to a value of 10. CorRet may also prove useful in other applications that discover the underlying ``topology'' (nearest-neighbor structure) of image collections. 
CorRet and CorLoc metrics effectively complement each other in the mixed-class setup: CorRet reveals how correctly an image is associated to other images, while CorLoc measures how correctly an object is localized in the image.

\subsection{The Object Discovery dataset}

The Object Discovery dataset~\cite{Rubinstein2013} was collected by the Bing API using queries for airplane, car, and horse, resulting in image sets containing outlier images without the query object. We use the 100 image subsets~\cite{Rubinstein2013} to enable comparisons to previous state of the art in cosegmentation and colocalization. In each set of 100 images, airplane, car, horse have 18, 11, 7 outlier images, respectively. Following ~\cite{Tang14}, we convert the ground-truth segmentations and cosegmentation results of~\cite{Kim2011,Joulin2010,Joulin2012,Rubinstein2013} to localization boxes. 

\begin{table}[t]
\caption{CorLoc (\%) on separate-class Object Discovery dataset.}
\begin{center}\vskip-0.3cm
\small
\addtolength{\tabcolsep}{-1.5pt}
\begin{tabular}{|c||c|c|c||c|}
\hline Methods & {Airplane} & {Car} & {Horse} & {Average} \\
\hline\hline
Kim \etal.~\cite{Kim2011} &  21.95 & 0.00 & 16.13 & 12.69\\
Joulin \etal.~\cite{Joulin2010} &  32.93 & 66.29 & 54.84 & 51.35\\
Joulin \etal.~\cite{Joulin2012} &  57.32 & 64.04 & 52.69 & 58.02\\
Rubinstein \etal.~\cite{Rubinstein2013} &  74.39 & 87.64 & 63.44 & 75.16\\
Tang \etal.~\cite{Tang14} &  71.95 & 93.26 & 64.52 & 76.58\\
Ours &  \textbf{82.93} & \textbf{94.38} & \textbf{75.27} & \textbf{84.19}\\
\hline
\end{tabular}
\vspace{-0.6cm}
\label{tab:OD_3a}
\end{center}
\end{table}
\begin{table}[t]
\caption{ Performance on mixed-class Object Discovery dataset.}
\begin{center}\vskip-0.3cm
\small
\addtolength{\tabcolsep}{-1.0pt}
\begin{tabular}{|c||c|c|c||c|}
\hline { Evaluation metric } & {Airplane} & {Car} & {Horse} & {Average} \\
\hline\hline
CorLoc  &  81.71 & 94.38 & 70.97 & 82.35\\
\hline
CorRet  &  73.30 & 92.00 & 82.80 & 82.70\\
\hline
\end{tabular}
\vspace{-0.65cm}
\label{tab:OD_3b}
\end{center}
\end{table}
\begin{figure}[t]
\centering
\begin{minipage}{0.495\linewidth}
\includegraphics[width=0.95\linewidth]{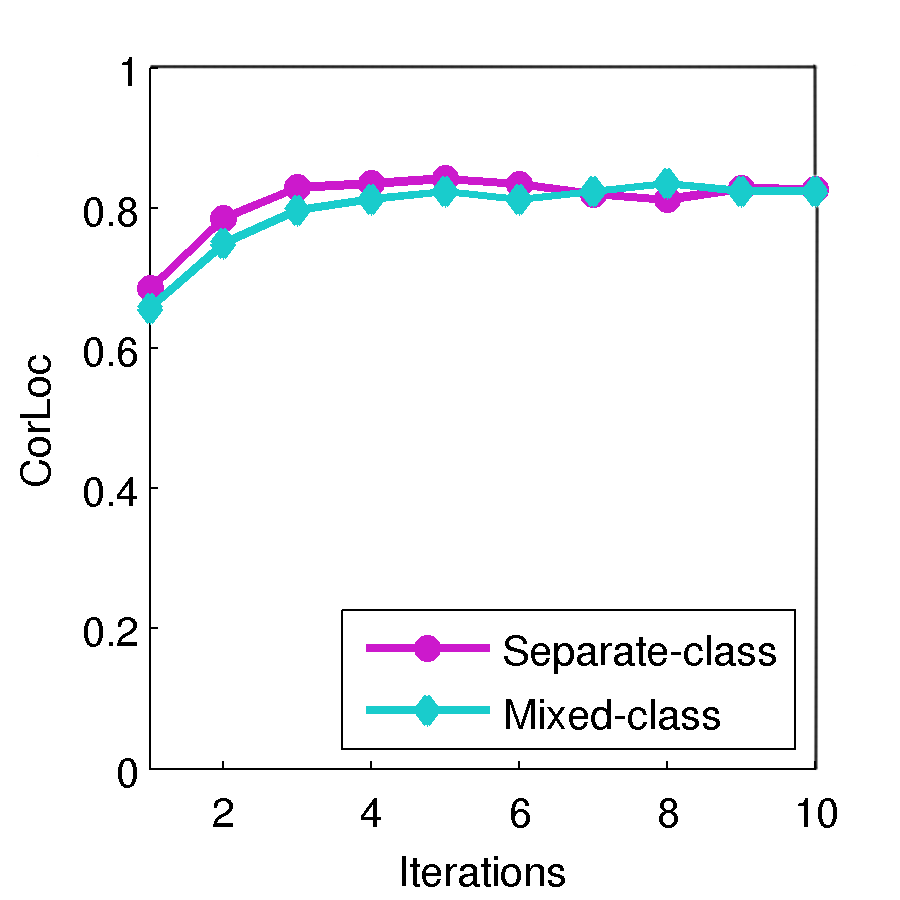}
\end{minipage}
\begin{minipage}{0.493\linewidth}
\includegraphics[width=0.95\linewidth]{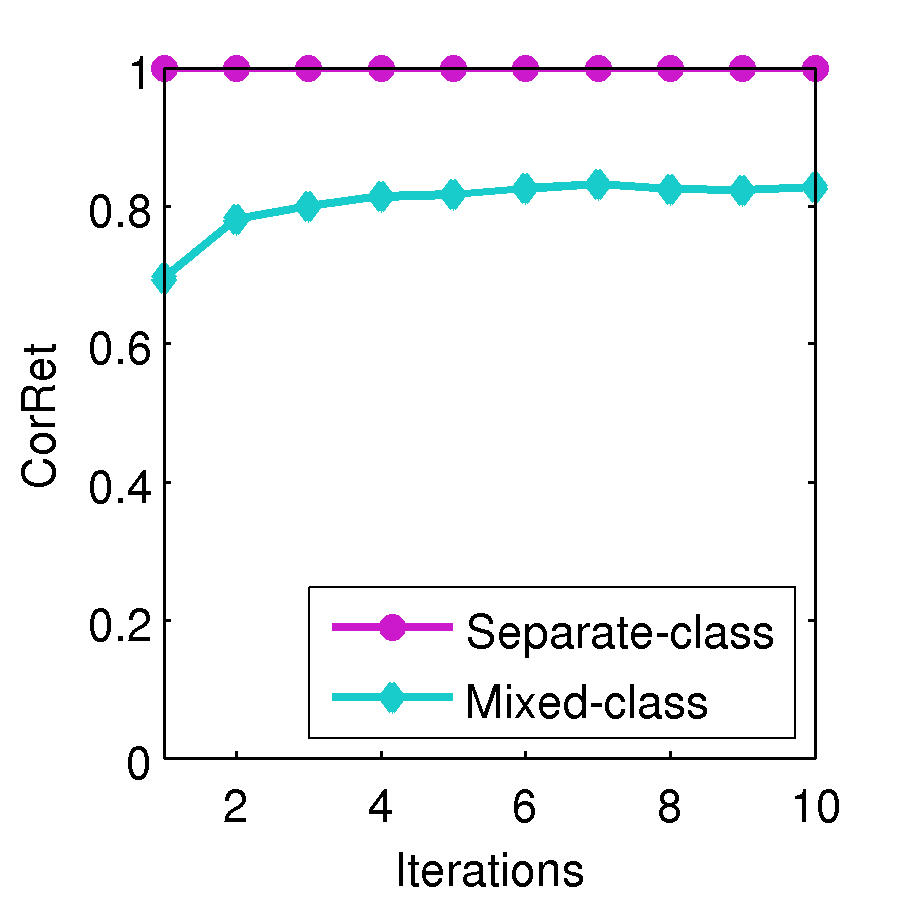}
\end{minipage}\vskip0.05cm
\caption{Average CorLoc (left) and CorRet (right) vs. \# of iterations on the Object Discovery dataset.}
\label{fig:exp_OD_iter}
\vspace{-0.2cm}
\end{figure}
\begin{figure}[t]
\centering
\begin{minipage}{1.0\linewidth}
\includegraphics[width=1.0\linewidth]{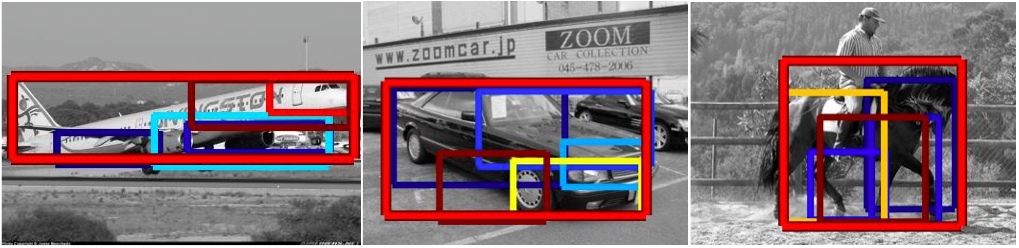}
\end{minipage}\vskip0.05cm
\caption{ Examples of localization on unlabeled Object Discovery dataset. Small boxes inside the localized object box (red box) represents five most confident part regions. (Best viewed in color.)  }
\label{fig:OD_3a}
\vspace{-0.3cm}
\end{figure}

We conduct separate-class experiments as in~\cite{Deselaers:2010he,Tang14}, and a mixed-class experiment on a collection of 300 images from all the three classes. The mixed-class image collection contains 3 classes and 36 outlier images. 
Figure~\ref{fig:exp_OD_iter} shows the average CorLoc and CorRet over iterations, where we see the proposed algorithm quickly improves both localization (CorLoc) and retrieval (CorRet) in early iterations, and then approaches a steady state. In the separate-class setup, CorRet is always perfect because no other object class exists in the retrieval. As we have found no significant change in both localization and retrieval after 4-5 iterations in all our experiments, we measure all performances of our method in this paper after 5 iterations.
The separate-class results are quantified in Table~\ref{tab:OD_3a}, and compared to those of state-of-the-art cosegmentation~\cite{Kim2011,Joulin2010,Joulin2012} and colocalization~\cite{Rubinstein2013,Tang14} methods. Our method outperforms all of them in this setup. The mixed-class result is in Table~\ref{tab:OD_3b}, and examples of localization are shown in Fig.~\ref{fig:OD_3a}. Remarkably, our localization performance in the mixed-class setup is almost the same as that in the separate-class setup. Localized objects are visualized in red boxes with five most confident regions inside the object, indicating parts most contributing to discovery.  Table~\ref{tab:OD_3b} and Fig.~\ref{fig:exp_OD_iter} show that our localization is robust to noisy neighbor images retrieved from different classes.    

\subsection{PASCAL VOC 2007 dataset}
The P{\small ASCAL} VOC 2007~\cite{pascal-voc-2007} contains realistic images of 20 object classes. Compared to the Object Discovery dataset, it is significantly more challenging due to considerable clutter, occlusion, and diverse viewpoints. To facilitate a scale-level analysis and comparison to other methods, we conduct experiments on two subsets of different sizes: P{\small ASCAL}07-6x2 and P{\small ASCAL}07-all.
\begin{table*}[t]
\caption{ CorLoc performance (\%) on separate-class P{\small ASCAL}07-6x2}
\begin{center}
\small
\addtolength{\tabcolsep}{-1.5pt}
\begin{tabular}{|c||c|c|c|c|c|c|c|c|c|c|c|c||c|}
\hline & \multicolumn{2}{|c|}{aeroplane} & \multicolumn{2}{|c|}{bicycle} & \multicolumn{2}{|c|}{boat} & \multicolumn{2}{|c|}{bus} & \multicolumn{2}{|c|}{horse} & \multicolumn{2}{|c||}{motorbike} & {  } \\
Method & L  & R & L & R & L & R & L & R & L & R & L & R & Average \\
\hline\hline
Ours (full) &  62.79 & 71.79 & 77.08 & 62.00 & 25.00 & 32.56 & 66.67 & 91.30 & 83.33  & 86.96  & 82.96   & 70.59  & \textbf{67.68}\\
 {\footnotesize Ours w/o MOR} &  62.79 & 74.36 & 52.08 & 42.00 & 15.91 & 27.91 & 61.90 & 91.30 & 85.42  & 76.09  & 48.72 & 8.82  & 53.94\\
 {\footnotesize Ours w/o PHM} &  39.53 & 38.46 & 54.17 & 60.00 & 6.82 & 9.30 & 42.86 & 73.91 & 68.75  & 82.61  & 33.33 & 2.94  & 42.72 \\
 {\footnotesize Ours w/o STO} &  34.88 & 0.0 & 2.08 & 0.0 & 0.0 & 4.65 & 0.0 & 8.70 & 64.58  & 30.43  & 2.56 & 0.0 & 12.32 \\
\hline
\end{tabular}
\vspace{-0.6cm}
\label{tab:exp_pascal07_6x2a}
\end{center}
\end{table*}
\begin{table*}[t]
\caption{ CorLoc and CorRet performance (\%) on mixed-class P{\small ASCAL}07-6x2.}
\begin{center}
\small
\addtolength{\tabcolsep}{-1.5pt}
\begin{tabular}{|c||c|c|c|c|c|c|c|c|c|c|c|c||c|}
\hline & \multicolumn{2}{|c|}{aeroplane} & \multicolumn{2}{|c|}{bicycle} & \multicolumn{2}{|c|}{boat} & \multicolumn{2}{|c|}{bus} & \multicolumn{2}{|c|}{horse} & \multicolumn{2}{|c||}{motorbike} & {  } \\
 {Metric} & L  & R & L & R & L & R & L & R & L & R & L & R & Average \\
\hline\hline
CorLoc &  62.79 & 66.67 & 54.17 & 56.00 & 18.18 & 18.60 & 42.86 & 69.57 & 70.83  & 71.74  & 69.23   & 44.12  & 53.73\\
\hline
CorRet &  61.40 & 42.56 & 48.75 & 56.80 & 19.09 & 13.02 & 13.33 & 30.87 & 41.46 & 41.74 & 38.72 & 43.24 & 37.58 \\
CorRet (class) &  \multicolumn{2}{|c|}{74.39} & \multicolumn{2}{|c|}{72.35} &\multicolumn{2}{|c|}{29.43} & \multicolumn{2}{|c|}{44.32} & \multicolumn{2}{|c|}{52.66} & \multicolumn{2}{|c||}{59.04} & 55.36 \\
\hline
\end{tabular}
\vspace{-0.65cm}
\label{tab:exp_pascal07_6x2b}
\end{center}
\end{table*}
\begin{table}[t]
\caption{ CorLoc comparison on P{\small ASCAL}07-6x2. }
\begin{center}
\small
\addtolength{\tabcolsep}{-1pt}
\begin{tabular}{|c|c|c|}
\hline
Method & Data used & Avg. CorLoc (\%) \\
\hline\hline
Chum and Zisserman~\cite{Chum2007} & P + N & 33 \\
Deselaers \etal.~\cite{Deselaers:2010he} & P + N & 50 \\
Siva and Xiang~\cite{siva2011weakly} & P + N & 49 \\
Tang \etal.~\cite{Tang14} & P & 39 \\
Ours &  P & \textbf{68}  \\
\hline\hline
Ours (mixed-class) & unsupervised &  54  \\
\hline
\end{tabular}
\vspace{-0.5cm}
\label{tab:exp_pascal07_6x2c}
\end{center}
\end{table}
\begin{figure}[t]
\centering
\begin{minipage}{0.16\linewidth}
\includegraphics[width=1.0\linewidth]{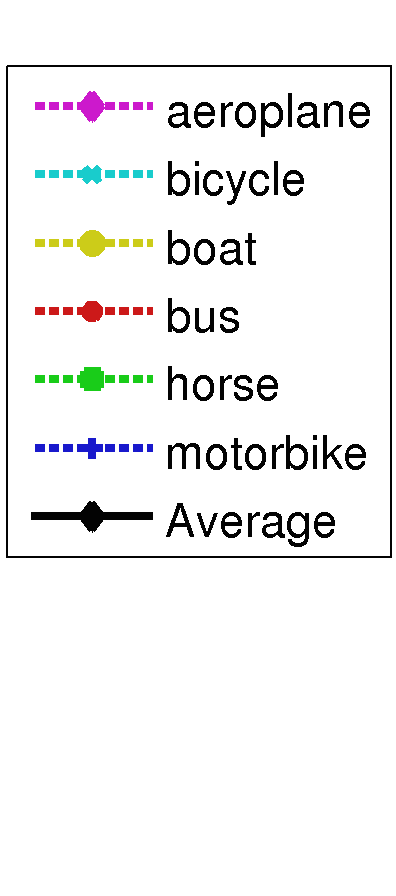}
\end{minipage}
\begin{minipage}{0.41\linewidth}
\includegraphics[width=1.0\linewidth]{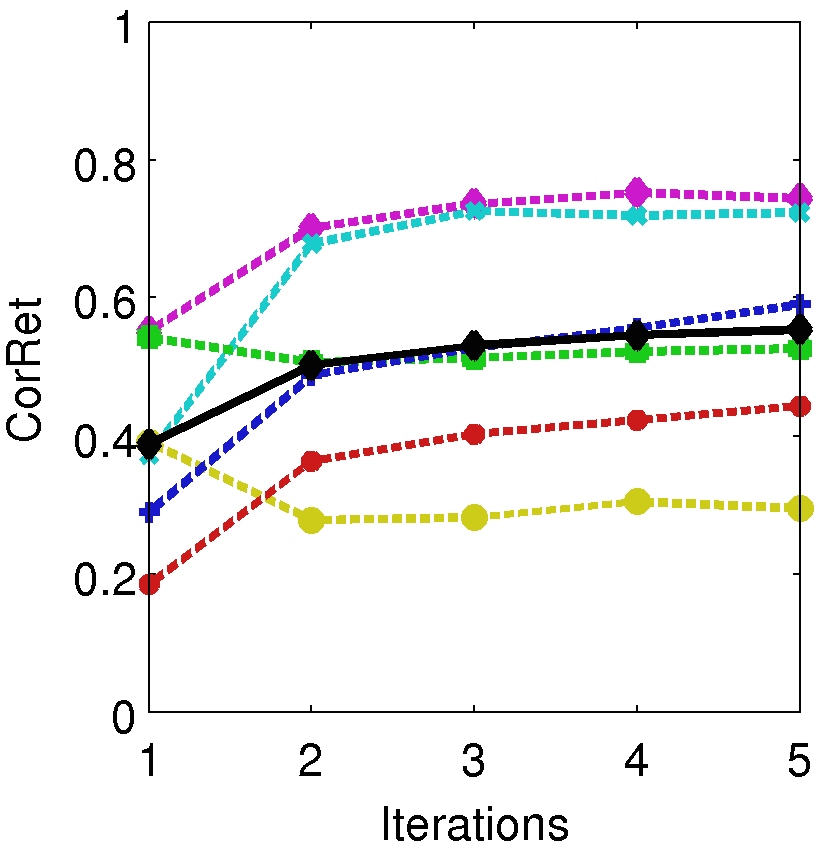}
\end{minipage}
\begin{minipage}{0.41\linewidth}
\includegraphics[width=1.0\linewidth]{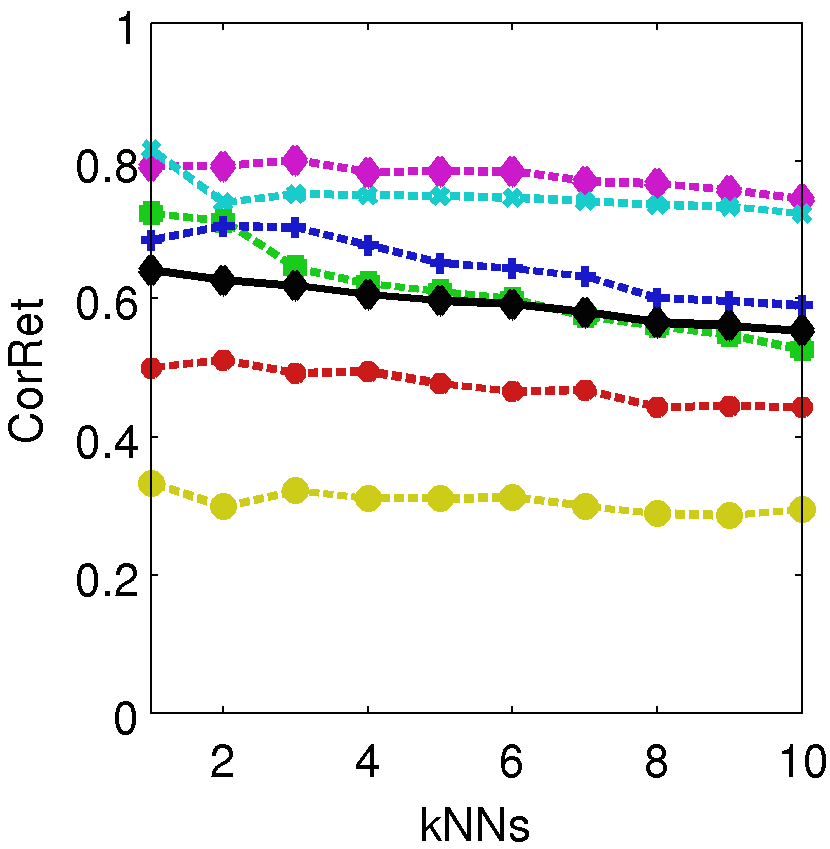}
\end{minipage}\vskip0.1cm
\caption{CorRet variation on mixed-class P{\small ASCAL}07-6x2.}
\label{fig:exp_pascal07_6x2_3_NN}
\vspace{-0.2cm}
\end{figure}
\begin{figure}[t]
\centering
\begin{minipage}{1.0\linewidth}
\includegraphics[width=1.0\linewidth]{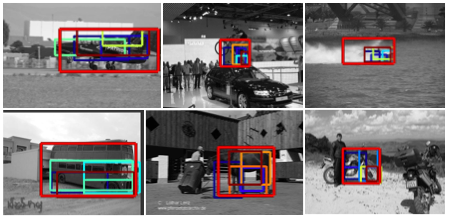}
\end{minipage}\vskip0.05cm
\caption{ Example results on mixed-class P{\small ASCAL}07-6x2.}
\label{fig:exp_pascal_6x2}
\vspace{-0.3cm}
\end{figure}
The P{\small ASCAL}07-6x2 subset~\cite{Deselaers:2010he} consists of all images from 6 classes (aeroplane, bicycle, boat, bus, horse, and motorbike) of train+val dataset from the left and right aspect each. Each of the 12 class/viewpoint combinations contains between 21 and 50 images for a total of 463 images. 
For a large-scale experiment with all classes following~\cite{Cinbis2014,Deselaers:2010he,Pandey2011}, we take all train+val dataset images discarding images that only contain object instances marked as ``difficult'' or ``truncate''. Each of the 20 classes contains between 49 and 1023 images for a total of 4548 images. 
We refer to it as P{\small ASCAL}07-all. 
\vspace{-0.4cm}
\paragraph{Experiments on P{\small ASCAL}07-6x2.}   
In the separate-class setup, we evaluate performance for each class in Table~\ref{tab:exp_pascal07_6x2a}, where we also analyze each component of our method by removing it from the full version: `w/o MOR' eliminates the use of multiple object regions over iterations, thus maintaining only a single potential object region for each image. `w/o PHM' substitutes PHM with appearance-based matching without any geometric consideration. `w/o STO' replaces the standout score with the maximum confidence. As expected, we can see that the removal of each component damages performance substantially. In particular, it clearly shows both part-based matching (using PHM) and foreground localization (using the standout score) are crucial for robust object discovery. 
In Table~\ref{tab:exp_pascal07_6x2c}, we quantitatively compare ours to previous results~\cite{Chum2007,Deselaers:2010he,siva2011weakly,Tang14} on P{\small ASCAL}07-6x2. Our method significantly outperforms those with a large margin. 
Note that our method does not incorporate any form of object priors such as off-the-shelf objectness measures~\cite{Deselaers:2010he,siva2011weakly,Tang14}, and only use positive images (P) without more training data,~\ie, negative images (N)~\cite{Deselaers:2010he,siva2011weakly}.  
For the mixed-class experiment, we run our method on a collection of all class/view images in P{\small ASCAL}07-6x2, and evaluate its CorLoc and CorRet peformance in Table~\ref{tab:exp_pascal07_6x2b}. To better understand our retrieval performance per class, we measure CorRet for classes (regardless of views) in the third row, and analyze it by increasing the numbers of iterations and neighbor images in Fig.~\ref{fig:exp_pascal07_6x2_3_NN}.   
This shows that our method achieves better localization and retrieval simultaneously, and benefits from each other.  
In Fig.~\ref{fig:exp_pascal_6x2}, we show example results of our mixed-class experiment on P{\small ASCAL}07-6x2. In spite of a small size of objects even partially occluded, our method is able to localize instances from cluttered scenes, and discovers confident object parts as well. From Table~\ref{tab:exp_pascal07_6x2c}, we see that even without using the separate-class setup, the method localizes target objects markedly better than recent colocalization methods.  
\vspace{-0.45cm}
\paragraph{Experiments on P{\small ASCAL}07-all.} 
Here we tackle a much more challenging and larger-scale discovery task, using all the images from the PASCAL07 dataset. We first report separate-class results, and compare our results to those of the state of the arts in weakly-supervised localization~\cite{Cinbis2014,Pandey2011,shi2013bayesian,siva2011weakly,siva2012defence,Siva2013,wang2014weakly} and colocalization~\cite{Joulin14} in Table~\ref{tab:exp_pascal07_all_separate}. 
Note that weakly-supervised methods use more training data,~\ie, negative images (N). Also note that the best performing  method~\cite{wang2014weakly} uses CNN features pretrained on the ImageNet dataset~\cite{imagenet_cvpr09}, thus additional supervised data (A). Surprisingly, the performance of our method is very close to the best of weakly-supervised localization~\cite{Cinbis2014} not using such additional data. 

In the mixed-class setting, we face an issue linked to the potential presence of multiple dominant labeled (ground-truth) objects in each image. 
Basically, both CorLoc and CorRet are defined as a per-image measure, \eg, CorLoc assigns an image true if any true localization is done in the image. For images with multiple class labels in the mixed-class setup, which is the case of P{\small ASCAL}-all with highly overlapping class labels (\eg, persons appear in almost 1/3 of images), CorLoc needs to be extended in a natural manner. 
To measure a class-specific average CorLoc in such a multi-label and mixed-class setup, we take all images containing the object class and measure their average CorLoc for the class. The upper bound of this class-specific average CorLoc may be less than 100\% because only one localization exists for each image in our setting. To complement this, as shown at the last column of Table~\ref{tab:exp_pascal07_all_mixed}, we add the `any'-class average CorLoc, where we assign an image true if any true localization of any class exists in the image. The similar evaluation is also done for CorRec. Both `any'-class CorLoc and CorRet have an upper bound of 100\% even when images have multiple class labels, whereas those in `Av.' (average) may not.
\begin{table*}[t]
\caption{ CorLoc (\%) on separate-class P{\small ASCAL}07-all, compared to the state of the arts in weakly-supervised / co- localization. }
\begin{center}
\footnotesize
\addtolength{\tabcolsep}{-4.15pt}
\begin{tabular}{|c|c||c c c c c c c c c c c c c c c c c c c c |c|}
\hline
{Method} & Data used & aero	& bicy & bird & boa & bot & bus & car & cat & cha & cow & dtab & dog & hors & mbik & pers & plnt & she & sofa & trai & tv & Av. \\
\hline\hline
Pandey \& Lazebnik~\cite{Pandey2011} & P + N & 50.9 & 56.7 & - & 10.6 & 0 & 56.6 & - & - & 2.5 & - &14.3 & - &50.0 &53.5 &11.2 & 5.0 &- &34.9 &33.0 & 40.6 & -\\
Siva \& Xiang~\cite{siva2011weakly} & P + N & 42.4 & 46.5 & 18.2 & 8.8 & 2.9 & 40.9 & 73.2 & 44.8 & 5.4 & 30.5 & 19.0 & 34.0 & 48.8 & 65.3 & 8.2 & 9.4 & 16.7 & 32.3 & 54.8 & 5.5 & 30.4 \\
Siva \etal.~\cite{siva2012defence} & P + N & 45.8 & 21.8 & 30.9 & 20.4 & 5.3 & 37.6 & 40.8 & 51.6 & 7.0 & 29.8 & 27.5 & 41.3 & 41.8 & 47.3 & 24.1 & 12.2 & 28.1 & 32.8 & 48.7 & 9.4 & 30.2 \\
Shi ~\etal.~\cite{shi2013bayesian} & P + N & 67.3 & 54.4 & 34.3 & 17.8 & 1.3 & 46.6 & 60.7 & 68.9 & 2.5 & 32.4 & 16.2 & 58.9 & 51.5 & 64.6 & 18.2 & 3.1 & 20.9 & 34.7 & 63.4 & 5.9 & 36.2 \\
Cinbis \etal.~\cite{Cinbis2014} & P + N & 56.6 & 58.3 & 28.4 & 20.7 & 6.8 & 54.9 & 69.1 & 20.8 & 9.2 & 50.5 & 10.2 & 29.0 & 58.0 & 64.9 & 36.7 & 18.7 & 56.5 &13.2 & 54.9 & 59.4 & 38.8 \\
Wang \etal.~\cite{wang2014weakly} & P + N + A & 80.1 & 63.9 & 51.5 & 14.9 & 21.0 & 55.7 & 74.2 & 43.5 & 26.2 & 53.4 & 16.3 & 56.7 & 58.3 & 69.5 & 14.1 & 38.3 & 58.8 & 47.2 & 49.1 & 60.9 &  48.5 \\
Joulin \etal.~\cite{Joulin14} & P & - & - & - & - & - & - & - & - & - & - & - & - & - & - & - & - & - &- & - & - & 24.6 \\
Ours & P & 50.3 & 42.8 & 30.0 & 18.5 & 4.0 & 62.3 & 64.5 & 42.5 & 8.6 & 49.0 & 12.2 & 44.0 & 64.1 & 57.2 & 15.3 & 9.4 & 30.9 & 34.0 & 61.6 & 31.5 & 36.6 \\
\hline
\end{tabular}
\vspace{-0.55cm}
\label{tab:exp_pascal07_all_separate}
\end{center}
\end{table*}
\begin{table*}[t]
\caption{ CorLoc and CorRet performance (\%) on mixed-class P{\small ASCAL}07-all. (See text for `any'). }
\begin{center}
\footnotesize
\addtolength{\tabcolsep}{-3.3pt}
\begin{tabular}{|c||c c c c c c c c c c c c c c c c c c c c |c||c|}
\hline
Evaluation metric & aero & bicy & bird & boa & bot & bus & car & cat & cha & cow & dtab & dog & hors & mbik & pers & plnt & she & sofa & trai & tv & Av. & any\\
\hline\hline
CorLoc & 40.4 & 32.8 & 28.8 & 22.7 & 2.8 & 48.4 & 58.7 & 41.0 & 9.8 & 32.0 & 10.2 & 41.9 & 51.9 & 43.3 & 13.0 & 10.6 & 32.4 & 30.2 & 52.7 & 21.8 & 31.3 & 37.6 \\
\hline
CorRet & 51.1 & 45.3 & 12.7 &  12.1 &  11.4 &  21.2  &  61.9  &  11.6 & 19.2 & 9.7 & 3.9 &  17.2 & 29.6 & 34.0  &  43.7 &  10.2  &  8.1  &  9.9 &  23.7  & 27.3  &  23.2 & 36.6 \\
\hline
\end{tabular}
\vspace{-0.7cm}
\label{tab:exp_pascal07_all_mixed}
\end{center}
\end{table*}
\begin{figure}[t]
\centering
\begin{minipage}{1.0\linewidth}
\includegraphics[width=1.0\linewidth]{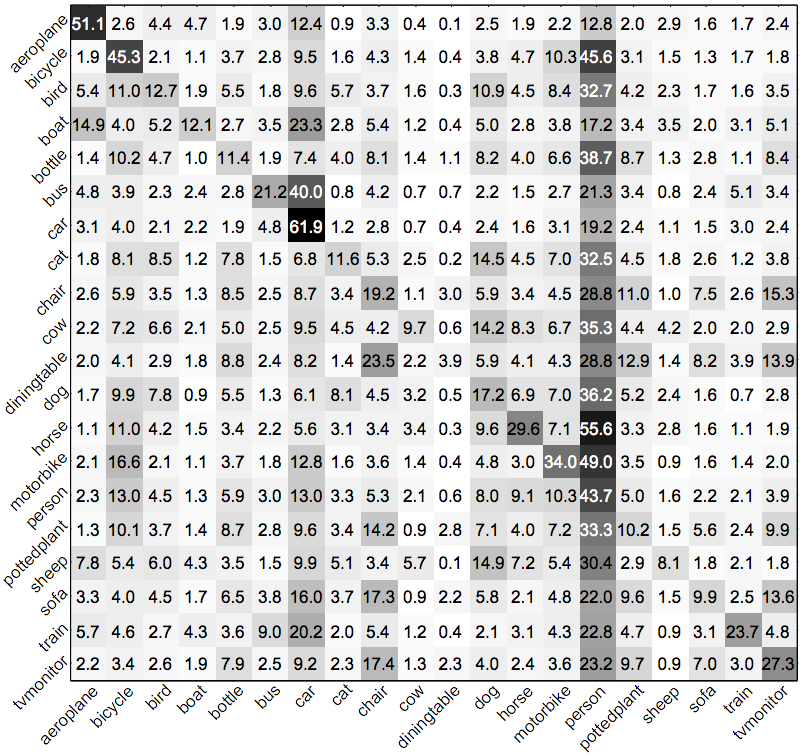}
\end{minipage}
\caption{ Confusion matrix of retrieval on mixed P{\small ASCAL}07-all. }
\label{fig:exp_pascal_all_confusion}
\vspace{-0.3cm}
\end{figure}
\begin{figure}[t]
\centering
\begin{minipage}{1.0\linewidth}
\includegraphics[width=1.0\linewidth]{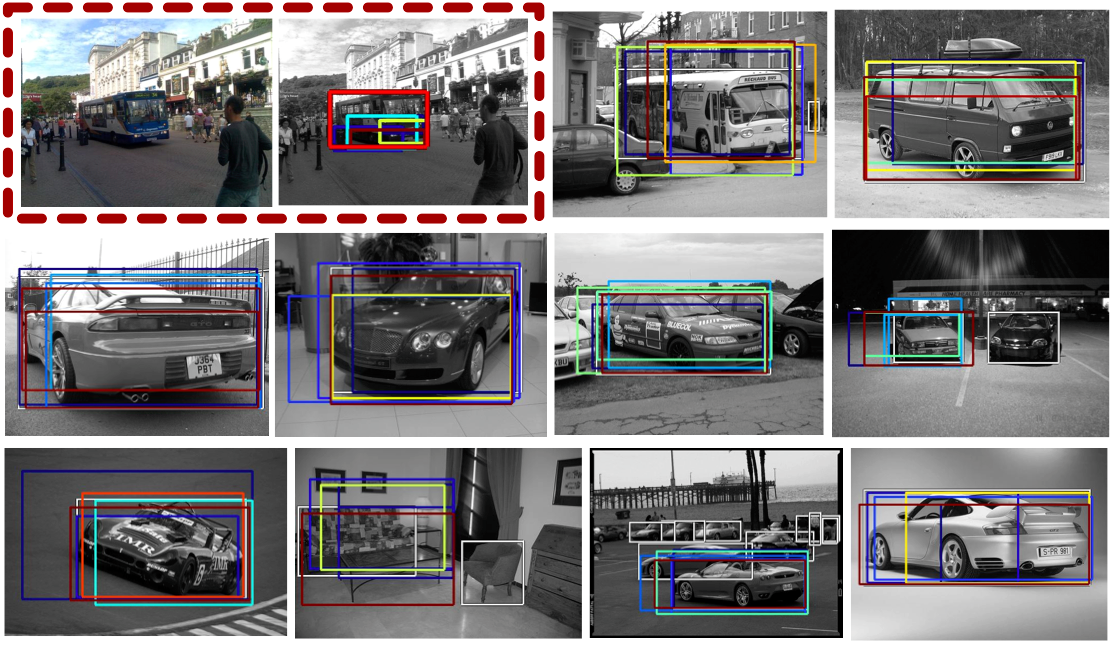}
\end{minipage}\vskip0.05cm
\caption{ Localization in an example and its neighbor images on mixed-class P{\small ASCAL}07-all. A bus is successfully localized in the image (red dashed box) from its neighbors (10 images) containing even other classes (car, sofa). Boxes in the neighbors show potential objects at the final iteration. (Best viewed in color.) }
\label{fig:exp_pascal_all_retrieval}
\vspace{-0.3cm}
\end{figure}
Note that the mixed-class PASCAL07-all dataset has a very imbalanced class distribution: the 20 classes have very different numbers of images, from 49 (sheep) to 1023 (person). Yet, as quantified in Table~\ref{tab:exp_pascal07_all_mixed}, our method still performs well even in this unsupervised mixed-class setting, and its localization performance is comparable to that in the separate-class setup. 
To better understand this, we visualize in Fig.~\ref{fig:exp_pascal_all_confusion} a confusion matrix of retrieved neighbor images based on the mixed-class result, where  
each row corresponds to the average retrieval ratios (\%) by each class.  Note that the matrix reflects class frequency so that the person class appears dominant. 
We clearly see that despite relatively low retrieval accuracy, many of retrieved images come from other classes with partial similarity, \eg, bicycle - motorbike, bus - car, etc. Figure~\ref{fig:exp_pascal_all_retrieval} shows a typical example of such cases. These results strongly suggest that our part-based approach to object discovery effectively benefits from different but similar classes without any class-specific supervision.    
Interestingly, the significant difference in retrieval performance (CorRet) from 100\% in the separate-class setup influences much less on localization (CorLoc). 
Further analysis of our experiments also reveals that in the case of an imbalanced distribution of classes, a class with lower frequency is harder to be localized than a class with higher frequency. To see this, consider `the highest' (person, car, chair, dog, cat) and `the lowest' (sheep, cow, boat, bus, dinningtable) in class frequency. We have measured how much the average performance changes between the separate-class (clean) and mixed-class (imbalanced) settings. The average CorLoc of `the highest' only drops by $1.2\%$, while that of `the lowest' drops by $9.4\%$. This clearly indicates that a class with lower class frequency is harder to localize in the mixed-class setting. Retrieval performance of `the lowest' (CorRet $11.0\%$) is also much worse than that of `the highest' (CorRet $30.7\%$). 
For more information, see our project webpage: \url{http://www.di.ens.fr/willow/research/objectdiscovery/}.   

\section{Discussion and conclusion}
We have demonstrated unsupervised object localization in the challenging mixed-class setup, which has never been fully attempted before on a challenging dataset such as~\cite{pascal-voc-2007}. 
The result shows that the effective use of part-based matching is a crucial factor for object discovery. 
In the future, we will advance this direction and further explore how to handle multiple object instances per image as well as build visual models for classification and detection. 
In this paper, our aim has been to evaluate our unsupervised algorithm per se, and have thus abstained from any form of additional supervision such as off-the-shelf saliency/objectness measures, negative data, and pretrained features. The use of such information will further improve our results. 

\vspace{-0.3cm}
\paragraph{Acknowledgments.}
This work was supported by the ERC grants Activia, Allegro, and VideoWorld, and the Institut Universitaire de France. 

{\small
\bibliographystyle{ieee}
\bibliography{minsucho_bib}
}

\end{document}